\documentclass[letterpaper, 10 pt, conference]{ieeeconf}  

\IEEEoverridecommandlockouts                              

\usepackage{graphics} 
\usepackage{epsfig} 
\usepackage{amsmath} 
\usepackage{amssymb}  
\usepackage{bm} 
\usepackage{multirow} 

\usepackage[flushmargin,hang]{footmisc} 

\newcommand{\fig}[1]{Fig.~\ref{#1}}

\def\epsgaiji#1{\leavevmode\kern-0.025zw\raise-.37zh\hbox{%
  \epsfile{file=#1,width=1.05zw}}\kern-0.025zw}
\newcommand{\MARU}[1]{{\ooalign{\hfil#1\/\hfil\crcr\raise.167ex\hbox{\mathhexbox20D}}}}

\usepackage{array}

\usepackage{flushend} 
\usepackage{comment}

\usepackage{siunitx} 
\usepackage{url}

\usepackage{pgfplots}
\pgfplotsset{compat=newest}
\usetikzlibrary{plotmarks}
\usetikzlibrary{arrows.meta}
\usepgfplotslibrary{patchplots}
\usepackage{booktabs} 
\usepackage{grffile}
\pgfplotsset{plot coordinates/math parser=false}
\newlength\fwidth
\newlength\fheight

\usepackage{cite}

\title{\LARGE \bf
Robot-Assisted Deployment and Maintenance of Inflatable Modules for Lunar Habitation: A Field Demonstration}

\author{Kentaro Uno$^{1\dag}$, Shamistan Karimov$^{1\dag}$, Ashutosh Mishra$^{1}$, Elian Neppel$^{1}$, Hazal Gozbasi$^{1}$, \\Shreya Santra$^{1}$, Shinichi Kimura$^{2}$, Kazuya Yoshida$^{1}$
\thanks{$^{*}$This work was supported by JST Moonshot R\&D Program, Grant Number JPMJMS223B.}%
\thanks{$^{1}$K. Uno, S. Karimov, A. Mishra, E. Neppel, H. Gozbasi, S. Santra, K. Yoshida are with the Space Robotics Lab. (SRL) in Department of Aerospace Engineering, Graduate School of Engineering, Tohoku University, Sendai 980-8579, Japan.}%
\thanks{$^{2}$S. Kimura is with Department of Electrical Engineering, Graduate School of Science and Technology, Tokyo University of Science, Noda 278-8510, Japan.}%
\thanks{
\textit{$^\dag$These authors contributed equally to this work. The corresponding author is Kentaro Uno.} (\tt{unoken@tohoku.ac.jp})
    }%
}%

\begin{document}

\maketitle
\thispagestyle{empty}
\pagestyle{empty}


\begin{abstract}
Long-term human habitation and in-situ development on the Moon open a new era of space utilization. In this context, robots are a key technology for facilitating the construction of future human outposts. Toward the deployment and establishment of human habitation modules on the lunar surface, we propose a combined system consisting of inflatable modules and a modular, reconfigurable robotic system.
This paper presents a report demonstrating various robot-assisted task executions using real hardware, namely the modular and reconfigurable robot MoonBot and the inflatable module HIDAS, to enhance the reliability of their deployment and maintenance.
The demonstrated tasks include robotic inspection during inflation, module position alignment, final safety locking, and three-dimensional mapping for post-deployment maintenance. All demonstrations were conducted either in a laboratory environment or at a lunar analogue test site. Finally, lessons learned are discussed to provide essential insights for this robotic application to future lunar habitation.
\end{abstract}

\section{Introduction}\label{introduction}
Recently, human interest in returning to the Moon has been growing, as exemplified by the recent successful crewed spacecraft flyby mission conducted under NASA’s Artemis program~\cite{artemis}. On the Moon, humans are expected to engage in scientific and creative activities that leverage the unique space environment.
Prior to the initiation of long-term habitation, robotics plays a vital role in terrain scouting, ground preparation, and preliminary infrastructure construction~\cite{Moon_base,sherwood2019principles}. Such foundational infrastructure comprises deployable components, including solar power generators, local and global communication systems, and shelters for radiation shielding and thermal protection, collectively forming a lunar human outpost.
\begin{figure}[t]
\centerline{\includegraphics[width=\linewidth]{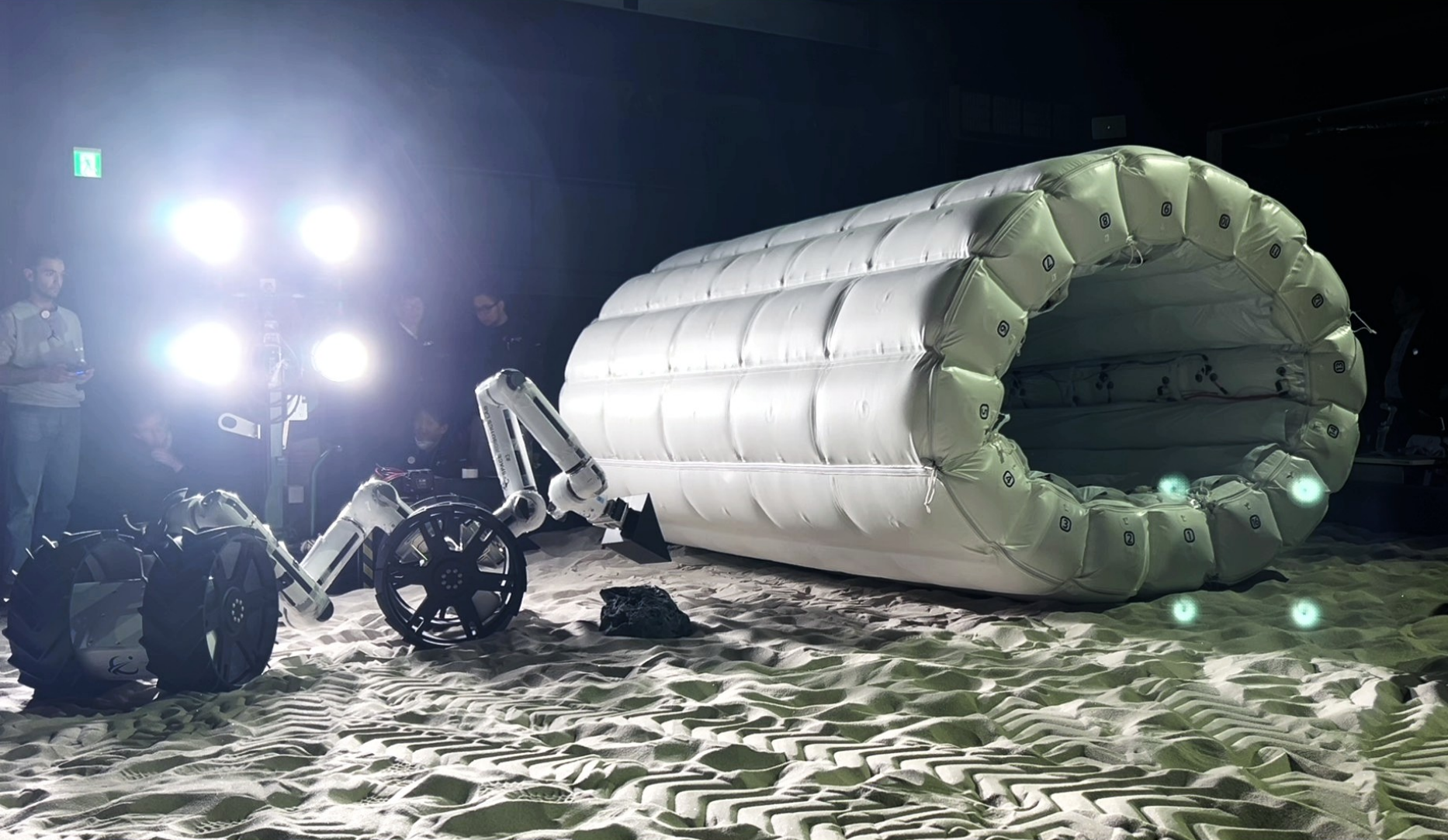}}
\vspace{-1mm}
\caption{MoonBot, a modular and reconfigurable robot (left), assisting in the deployment of inflatable module HIDAS (right). For the Earth-gravity demonstration, a half-scale inflatable module (diameter: 2\;m) was used. The robots monitor the inflation status using a hand-mounted camera, assist in positioning the module by applying external forces, and subsequently insert rolling-proof stoppers to prevent unintended movement.
}
\label{fig1}
\end{figure}
To efficiently construct such a lunar outpost, a practical approach is to launch components as modular units and assemble them in situ, similar to previous large-scale space constructions such as the International Space Station (ISS). One of the major advantages of this module-by-module construction approach is its ability to incrementally expand and customize the system in response to evolving mission requirements~\cite{gregg2024ultralight,uno2025moonbot,netti2025planetary}. In addition, such reconfigurable systems can be flexibly adapted to accommodate changes in mission objectives and environmental conditions.

Our proposal involves the same concept. Nonetheless, we aim to develop a modular and reconfigurable robotic system, similar to those proposed in~\cite{hancher2006modular,lordos2022worms}, while ensuring that the assembled infrastructure components are also modular, thereby enabling consistent modularity across both robotic systems and infrastructure components. The combination of such heterogeneous modules---where both the assembling robotic system and the assembled infrastructure are modular---provides enhanced adaptability, allowing the system to flexibly reconfigure in response to site-specific conditions that may only become apparent upon deployment (see \fig{fig1}).
\subsection{Related Work}
Various approaches have been studied for constructing habitat modules on the Moon and Mars. For instance, structures can be built using blocks or walls fabricated from lunar sand: regolith---blocks produced by sintering~\cite{caluk2023introduction} or walls created via additive manufacturing~\cite{bao2024situ,yim2024role}. Alternatively, natural terrain features, such as lava tubes, can be repurposed for habitation~\cite{boston2010location,sauro2020lava}. Accordingly, robotic exploration within such environments has attracted significant research interest~\cite{dominguez2025cooperative,chen2024locomotion,kolvenbach2025lunarleaper}.
Although leveraging in-situ resources is an effective way to reduce transportation costs, significant technological gaps remain before these approaches can be realized as practical solutions.

\begin{figure}[t]
    \centerline{\includegraphics[width=\linewidth]{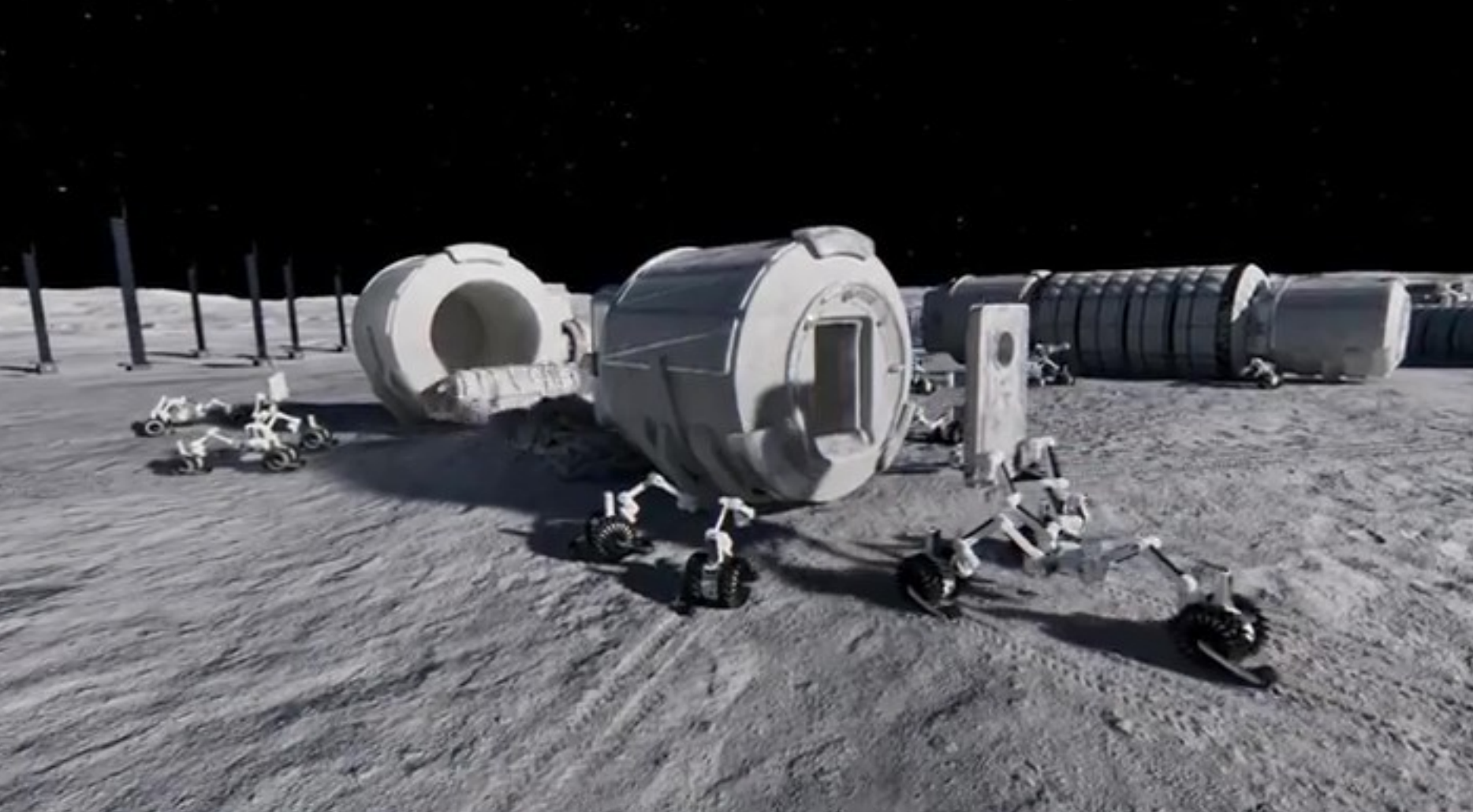}}
    \vspace{-1mm}
    \caption{Conceptual rendering illustrating the assumed scenario of robot-assisted construction of a human habitation base.
\label{fig2}}
\end{figure}

Therefore, as a short- to mid-term solution, inflatable modules have attracted considerable attention as a viable option for habitat construction. Such modules can be launched in an uninflated state, enabling compact packaging. In addition, the vacuum environment on the Moon reduces the amount of air required for inflation, which further helps to decrease the launch mass~\cite{roberts1992inflatable,qi2021review,Kimura_HIDAS}.
For the reliable deployment and sustainable maintenance of the module, robotic assistance is essential. In this regard, robots are required to monitor the inflation process, adjust positioning, attach additional exterior components, and perform safety locking (e.g., ground anchoring), followed by regular inspection and maintenance routines, as illustrated in \fig{fig2}.

\subsection{Contribution}
Numerous studies have investigated the use of inflatable systems as habitats for humans on the Moon. However, most of these studies have remained at the conceptual level rather than being experimentally demonstrated in realistic and practical scenarios. This paper presents dedicated mission-oriented demonstrations in laboratory and lunar analogue environments using real-world hardware. Rather than demonstrating inflatable habitats themselves, this study investigates how modular mobile robots can reliably assist their deployment and maintenance through representative mission tasks, including inspection, positioning, stabilization, and post-deployment mapping. Lessons learned throughout the field tests help address the current lack of knowledge regarding robotic operation of inflatable habitat modules for future lunar missions.

\section{Method}
To verify the effectiveness of the concept of robot-assisted deployment of infrastructural modules, dedicated demonstrations using real hardware were conducted. For this purpose, a modular robotic system called MoonBot and inflatable modules named HIDAS were prepared (see \fig{fig1}). In addition, we developed a human-in-the-loop teleoperation software in which the operator seamlessly switches between morphology-specific controllers and performs safer decision-making based on the shared autonomy concept. This section details the hardware and software used in this work.
\subsection{Hardware}
To demonstrate the secure and reliable deployment and maintenance of infrastructural units, such as human habitation modules, we employ MoonBot~\cite{uno2025moonbot}, a modular and reconfigurable robotic system designed for lunar applications. The system consists of multiple types of functionally minimal modular units, including the {\it Limb} module: an articulated unit for manipulation, and the {\it Wheel} module: a dual-wheeled unit for locomotion. The latest version of MoonBot's Limb module is a symmetric seven-degree-of-freedom (7-DOF) manipulator with a length of 1.2\;m and a mass of 15\;kg, equipped with a gripper and a camera at both ends. The Wheel module, with a mass of 28\;kg, serves as a mobile base with two independently driven, grousered 0.5\;m diameter wheels and includes grapple fixtures that enable connection with Limb modules, allowing serial limb--wheel configurations.
This self-reconfigurable modular robot can vary its structure and degrees of freedom to adapt to diverse tasks, such as carrying, manipulating, and assembling components.

For human habitation, we consider inflatable structures as the primary shelter for lunar base activities. These structures are designed to protect robots and humans from hazards such as cosmic radiation and extreme temperature variations. In this study, we employ HIDAS (Homeostatic Inflatable Decentralized Autonomous Structures)~\cite{Kimura_HIDAS}, a full-scale prototype of this inflatable module concept. HIDAS consists of multiple decentralized, pressurizable cells arranged in a cylindrical configuration. By controlling the internal pressure of these cells, the module can traverse the ground by rotating its whole body as part of a modular robotic system~\cite{HIDAS_mobility}.

\begin{figure}[t]
    \centering
    \includegraphics[width=.95\linewidth]{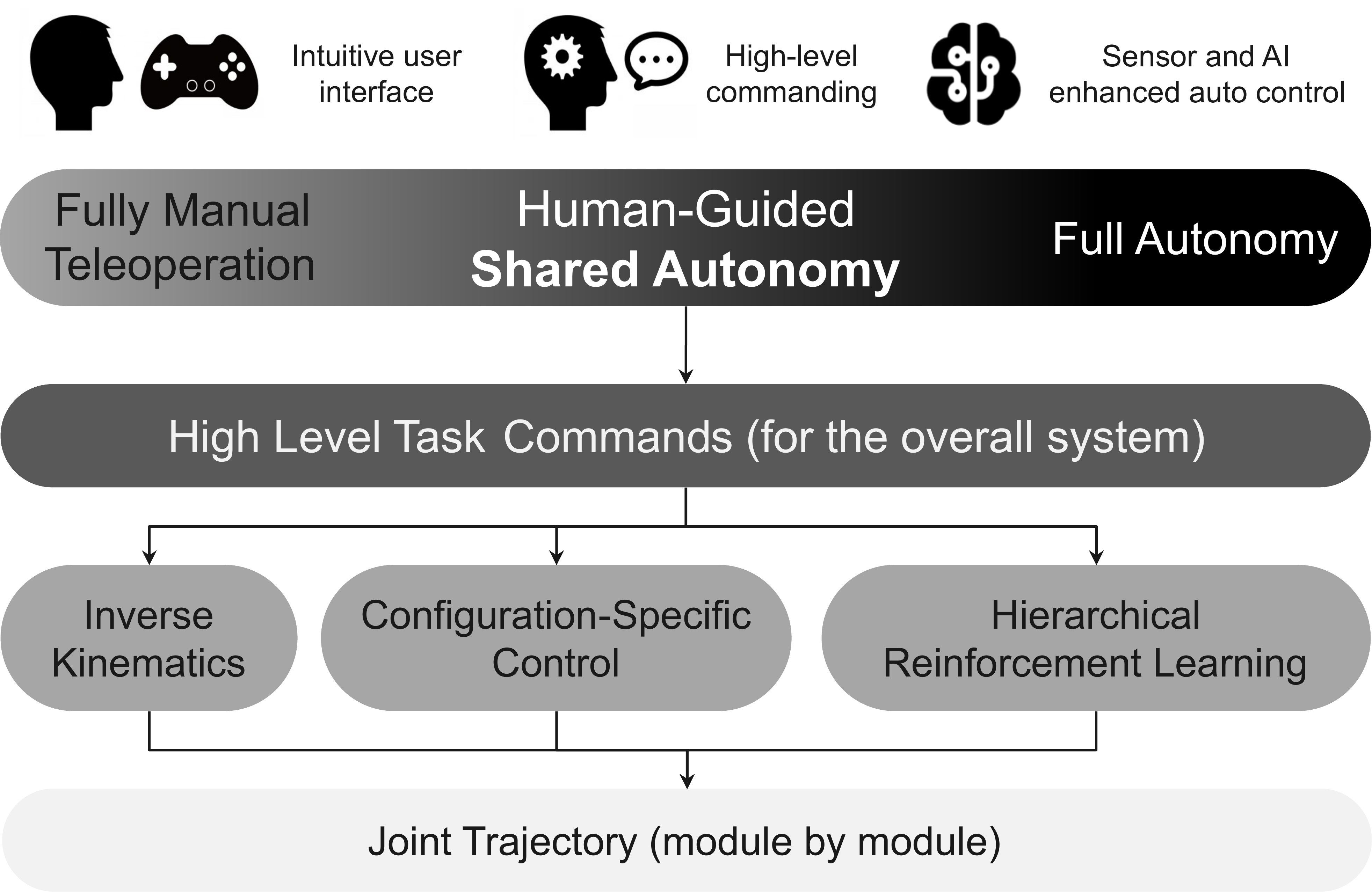}
    \vspace{-1mm}
    \caption{Concept of the robot teleoperation framework based on shared autonomy.
\label{software_concept}}
\end{figure}
HIDAS has a two-layer structure consisting of an outer shell that provides structural strength and enhanced resistance to external mechanical loads, and an inner bladder that stores the pressurized air. The outer shell is made of polyvinyl chloride (PVC)-coated polyester fabric with a thickness of 0.5\;mm, while the inner bladder is made of polyurethane film with a thickness of 0.2\;mm. While the full-scale HIDAS module is designed with a fully inflated diameter of 4\;m and a length of 6\;m, in this paper, a half-scale model (length: 3\;m, diameter: 2\;m, weight: 100\;kg) was used for the field demonstrations under Earth gravity.

\subsection{Software}
\begin{figure}[t]
    \centering
    \includegraphics[width=\linewidth]{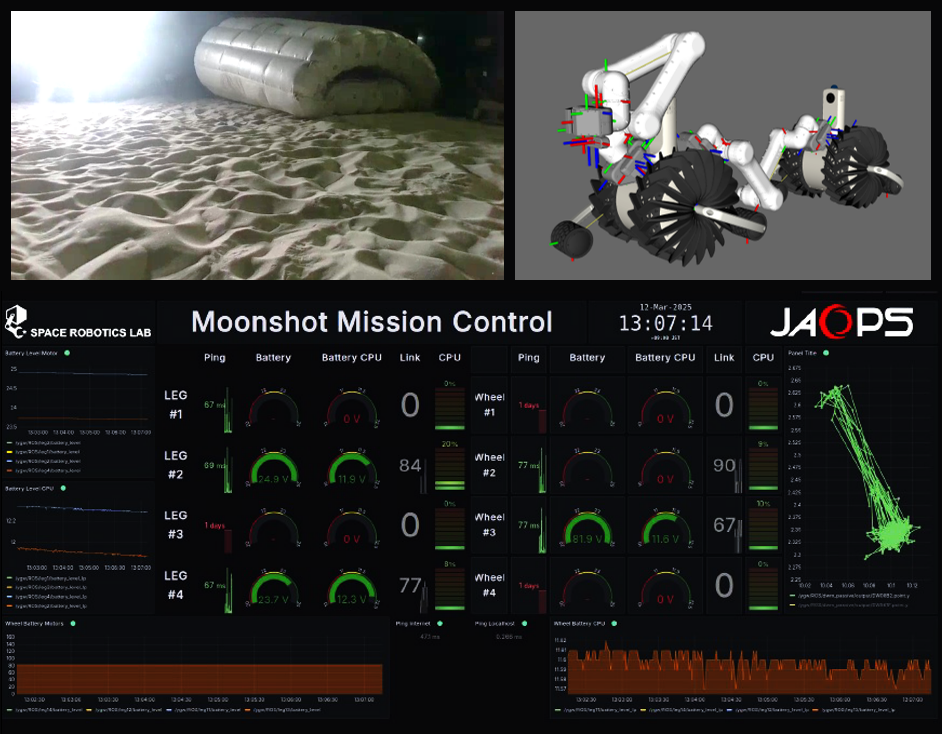}
    \vspace{-5mm}
    \caption{Robot teleoperation user interface. Top left: Robot view monitor, top right: joint states visualizer, bottom: telemetry data monitor (active modules, battery life, etc.).
\label{ui}}
\end{figure}
Space missions require high reliability; therefore,
human-in-the-loop operation remains essential, especially for
contact-rich tasks such as object lifting, handling, and
component insertion. However, direct joint-level teleoperation
of a modular robot is inefficient because the controllable
degrees-of-freedom change with the connected module
configuration. We therefore developed a shared-autonomy
teleoperation framework that lets the operator command the
robot at a task level, while the software handles
configuration-dependent motion generation, actuator
synchronization, and safety-aware command limitation
(see \fig{software_concept}).

The software architecture consists of an operator interface
layer, a configuration-aware control layer, and a low-level
synchronization layer. Through the operator interface, the
human operator sends high-level commands, such as
end-effector pose increments, gripper commands, and wheel
velocity commands, while direct joint-level commands remain
available for calibration and fine adjustment. The interface
also provides live camera feedback, joint-state visualization,
module status, battery state, and telemetry information,
allowing the operator to supervise the task and select the
appropriate controller for the current robot morphology
(\fig{ui}).

In the configuration-aware control layer, operator commands
are mapped to the corresponding controller according to the
current Limb---Wheel connection. For manipulation, Cartesian
end-effector commands are converted into joint-space
references through inverse kinematics. For locomotion and
pushing, wheel velocity commands are generated, while
manipulator commands are used for fine positioning or contact
adjustment. Thus, the operator can control the overall
loco-manipulation behavior without manually considering the
full kinematic structure of each reconfigured system.

At the low level, module-wise commands are executed through
{\it Motion Stack}~\cite{neppel2025robust}, a hardware-agnostic
actuator synchronization framework that provides a unified
command interface for multiple onboard controllers and
synchronizes joint trajectories across modules. Together with
the distributed MoonBot software architecture~\cite{neppel2025designing},
this enables each module to be addressed, monitored, and
controlled as part of a reconfigurable robot system.

To improve stability during contact-rich tasks, we incorporate
an end-effector alignment controller with adaptive velocity
bounds based on a dynamic hypersphere clamp, which was introduced and quantitatively evaluated in our previous work~\cite{karimov2025agnostic}.
Translational and rotational velocity commands are jointly
limited in a normalized six-dimensional command space and
scaled back when they exceed the allowable bound. This
prevents abrupt motion and improves robustness during
alignment, pushing, and insertion tasks.
\begin{table}[b]
  \centering
  \footnotesize
  \caption{Terminal alignment residuals from repeated trials.}
  \vspace{-2mm}
  \label{tab:controller_validation}
  \begin{tabular}{l|rrrr}
  \toprule
  Metric & Take 1 & Take 2 & Take 3 & Mean $\pm$ s.d. \\
  \midrule
  $\Delta d$ [mm] & 3.72 & 4.52 & 5.29 & $4.51 \pm 0.64$ \\
  $\Delta\theta$ [deg] & 0.55 & 0.53 & 0.69 & $0.59 \pm 0.07$ \\
  \bottomrule
  \end{tabular}
\end{table}
\begin{figure}[t]
    \centering
    \includegraphics[width=\linewidth]{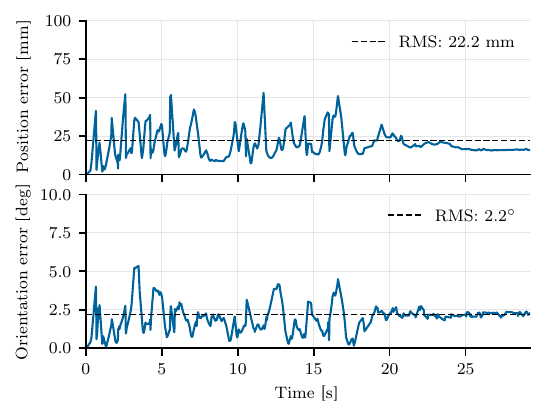}
    \vspace{-7mm}
    \caption{Time histories of the commanded-versus-measured hand-pose errors
    in the preliminary controller experiment performed in the lab. Dashed
    lines indicate the RMS errors.}
    \label{controller_tracking}
\end{figure}

Task-space performance was evaluated with an independent 16-camera OptiTrack
system in the experiment performed in our previous work~\cite{karimov2025agnostic}. Table~\ref{tab:controller_validation} summarizes the terminal
alignment residuals from three repeated takes, with mean translational and
rotational errors of $4.51 \pm 0.64$~mm and $0.59 \pm 0.07^\circ$,
respectively. For the response analysis, \fig{controller_tracking} compares
the Cartesian hand-pose reference with the OptiTrack measurement from Take~2
of the same experiment~\cite{karimov2025agnostic}. After transforming the
reference into the OptiTrack frame and compensating for the fixed marker
offset, the position error is computed as the 3-D Euclidean distance and the
orientation error as the quaternion angular difference. Their time histories
have root-mean-square (RMS) values of 22.2~mm and $2.2^\circ$, respectively. This tracking
response is representative of the continuous pose adjustments used in manual
HIDAS teleoperation. The HIDAS demonstrations themselves were not equipped
with external motion capture.

The framework also allows the operator to switch from manual
teleoperation to more autonomous behaviors when appropriate.
For example, sensor- and AI-enhanced functions, such as visual
anomaly detection and hierarchical reinforcement
learning-based motion policy generation, can be selected as
assistive controllers. In the hierarchical reinforcement learning
approach, module-level policies are fused according to the
current robot connection structure~\cite{mishra2025multi,minamikawa2026motion,mishra2026distributed}.
Thus, the developed software provides a unified interface for heterogeneous modular robot operation without
assuming a fixed robot morphology.

\section{Field Demonstration}\label{field-demo}
This paper serves as an initial report demonstrating the proof of concept of robot-assisted operations for the reliable deployment of inflatable modules. In particular, the objective of the demonstrations is to validate the feasibility and effectiveness of robotic assistance throughout the deployment process of inflatable habitat modules. To this end, the demonstrations were designed to cover the representative robotic tasks required for the entire deployment workflow of inflatable habitat modules, including inflation inspection, module positioning and alignment, final stabilization, and post-deployment inspection.

The task-level goals were defined according to the required physical outcome of each operation. For instance, to properly position the inflatable module, the robots were required to rotate or translate HIDAS until its longitudinal axis and end ring reached the intended relative alignment with the adjacent module. To stabilize the inflatable module for safety locking, the robot had to place the stopper beneath an end ring of HIDAS and the stopper had to remain in contact after release by the robot. The operators evaluated these outcomes using the live camera views and robot telemetry. Because no external tracking system or fixed global pose reference was installed during these trials, the present demonstrations quantify task completion and operation time rather than task-specific pose accuracy; independently measured controller accuracy is reported in the Software subsection.

A field testing campaign was conducted in a dedicated lunar analogue test field, while some demonstrations were performed in a laboratory environment. The lunar analogue field tests were carried out at the Space Exploration Field, a \SI{20}{m} $\times$ \SI{20}{m} experimental site fully covered with silica type sand and located within the Advanced Facility for Space Exploration at the Japan Aerospace Exploration Agency (JAXA).

\subsection{Inflation inspection}
\begin{figure}[t]
    \footnotesize
    \centering
    \includegraphics[width=\linewidth]{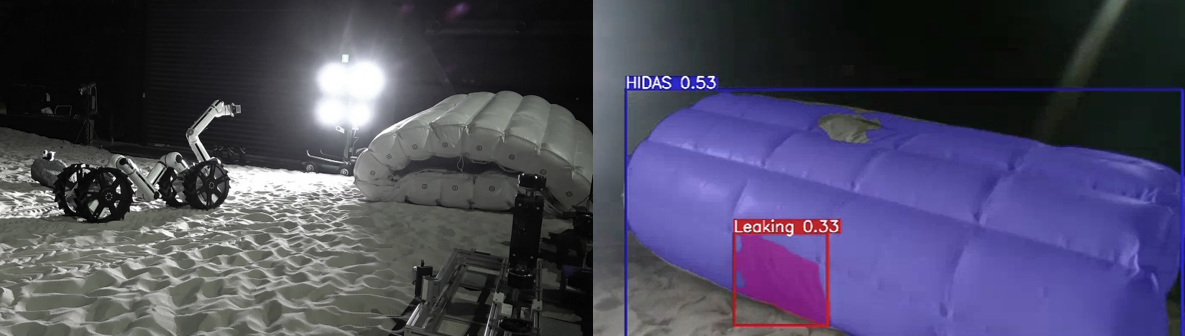} \\
    (a) Vision-based inspection. \\
    \vspace{2mm}
    \raggedright
    \includegraphics[width=.78\linewidth]{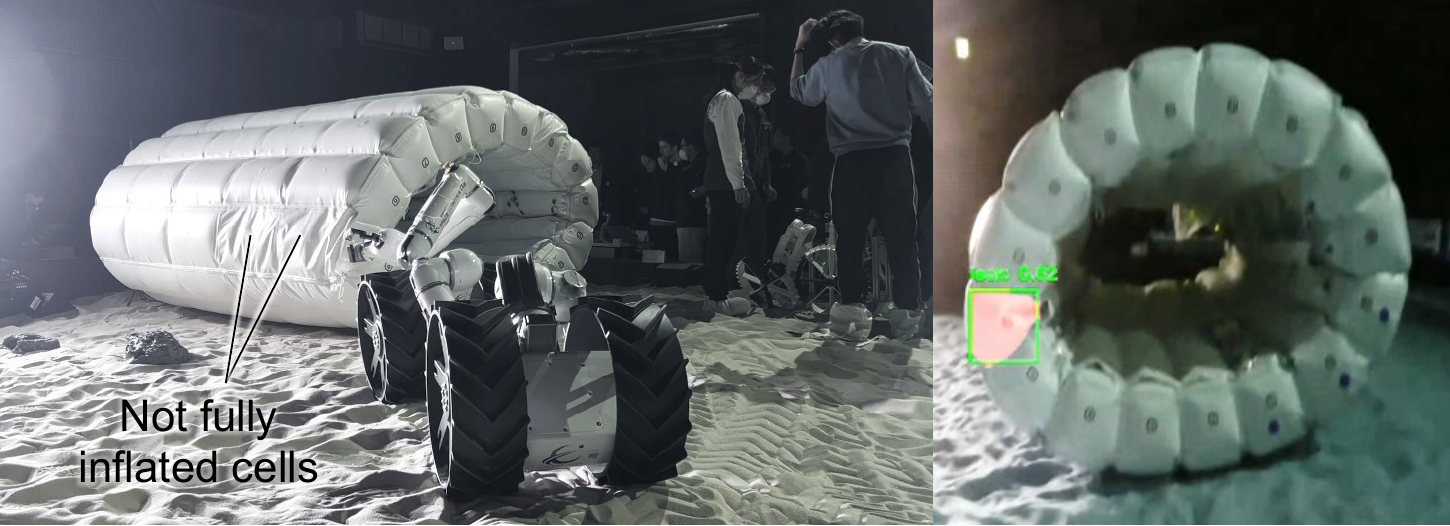} \\
    \centering
    (b) Contact-based inspection.
    \caption{Initial inspection of the inflatable module by the robot: (a) Vision-based inspection during inflation and (b) Contact-based inspection after inflation.
\label{initial_inspection}}
\end{figure}
During inflation, it is essential to monitor the process to ensure the absence of severe membrane damage. In the demonstration, a half-scale HIDAS prototype was placed in the sand field. An initial version of MoonBot, whose hardware specifications are detailed in \cite{uno2025moonbot}, was formed in the {\it Dragon} configuration (2 × Limbs + 2 × Wheels serially connected), enabling stable locomotion and manipulation, and was used for this demonstration.
\begin{table}[b]
  \centering
  \footnotesize
  \caption{Cell-inspection model (YOLOv11n-seg) performance on the held-out validation split. ``All'' is the unweighted mean over the two classes.}
  \vspace{-2mm}
  \label{tab:yolo}
  \begin{tabular}{l|cccc}
  \toprule
  Class & Precision & Recall & mAP@50 & mAP@50:95 \\
  \midrule
  HIDAS  & 0.995 & 0.991 & 0.995 & 0.988 \\
  Anomaly cell  & 0.765 & 0.779 & 0.793 & 0.551 \\
  \midrule
  All             & 0.880 & 0.885 & 0.894 & 0.770 \\
  \bottomrule
  \end{tabular}
  \end{table}
\begin{figure*}[t]
    \footnotesize
    \centering
    \centerline{\includegraphics[width=\linewidth]{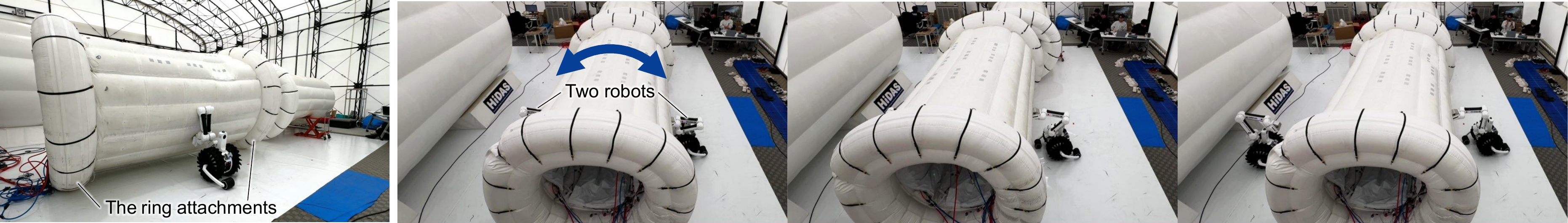}}
    (a) Lateral alignment. \vspace{2mm} \\
    \centerline{\includegraphics[width=\linewidth]{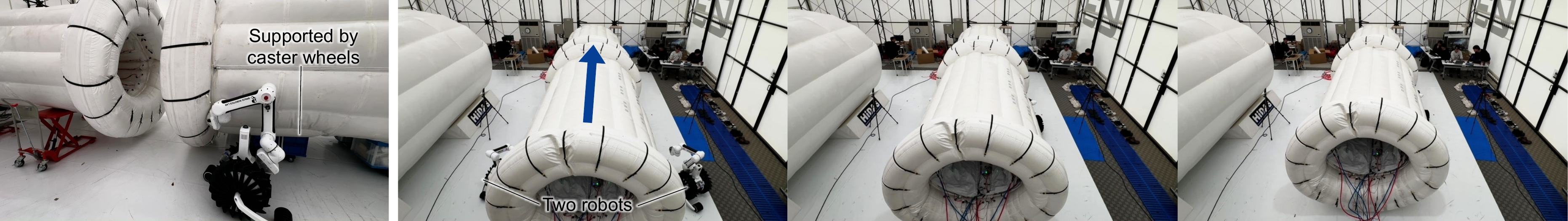}}
    (b) Longitudinal alignment.
    \caption{Well-positioning of the inflated module by two mobile robots, assuming a scenario in which the module is docked to another module. (a) The lateral position was adjusted by applying rolling forces from the left and right sides. Subsequently, (b) the longitudinal position was aligned by pushing the module cooperatively using the two robots.
\label{well-positioning}}
\end{figure*}

First, MoonBot looked around using a hand-mounted camera (RealSense D455), autonomously recognized HIDAS, and drove toward it. Subsequently, during inflation, the operator manually teleoperated the front manipulator of MoonBot Dragon using a simple inverse kinematics controller to inspect the HIDAS cells using the hand-eye camera to confirm the absence of hardware problems. For the autonomous detection of the HIDAS body and cells that were not inflating properly, which may indicate gas leakage, a YOLOv11~\cite{yolov11}-based instance-segmentation system was implemented. Because the optical environment under lunar illumination is highly challenging, characterized by a high dynamic range caused by direct, parallel sunlight in the absence of an atmosphere~\cite{camille_SII}, we reproduced similar illumination conditions using a halogen lamp to create a dedicated dataset for training. In total, 1,964 images were annotated and divided into training, validation, and test sets following a 70 : 20 : 10 ratio. All images were auto-oriented and resized to $640 \times 640$, while the training set was further expanded through horizontal and vertical flip augmentation. The custom-tuned YOLOv11n-seg model was trained for up to 300 epochs and evaluated on the held-out validation split using the standard instance-segmentation protocol, i.e., mask-level precision, recall, and mean average precision at a mask IoU (Intersection over Union) threshold of 0.50 (mAP@50) and averaged over IoU thresholds from 0.50 to 0.95 (mAP@50:95). As summarized in Table~\ref{tab:yolo}, the model segmented under-inflated cells with an mAP@50 of 79.3\% and an mAP@50:95 of 55.1\%, while the class-averaged mAP@50 over both classes reached 89.4\%, successfully identifying cells that were not sufficiently inflated (see top in \fig{initial_inspection}).

In addition, after inflation was completed, the operator further controlled the robot to perform a haptic inspection of HIDAS by sequentially applying a pushing force cell by cell, relying on the monitor screen showing the position of the automatically detected anomaly cells, each of which was equipped with an independent internal pressure sensor. By monitoring the increase in internal pressure in response to the applied external force, the system verified the completion of inflation (see the bottom of \fig{initial_inspection}). It is worth mentioning that no damage was observed on the outer shell of HIDAS during contact with the aluminium robot gripper.

As a result, the developed vision-based inspection system successfully detected anomalies during inflation, and the additional haptic inspection further improved reliability, demonstrating an effective initial inspection method for inflatable module deployment.

\subsection{Well-positioning}
HIDAS has locomotion capabilities by controlling the inflation level of each cell, enabling lateral rolling motion and longitudinal ``inchworm-like'' motion. However, precise self-positioning remains challenging under sandy terrain conditions on the Moon; therefore, robot-assisted well-positioning and final alignment are essential. To demonstrate this, two MoonBots in the {\it Minimal} configuration (1 × Limb + 1 × Wheel, docked) were operated. The robots worked collaboratively to push HIDAS and align it with another module, assuming the expansion of the human habitation space. These demonstrations were conducted by human operators teleoperating the robots from a hidden area in the laboratory environment.

First, the lateral position was aligned by rotating HIDAS using MoonBot Minimals from both the left and right sides. In these demonstrations, HIDAS was equipped with a ring-shaped inflatable attachment at both edges to enhance stability. The ring was inflated to a high pressure, enabling the main module to better maintain its cylindrical shape and reduce friction during rotation by minimizing the contact area with the ground. The robot pushed the inflatable module through wheel-driven motion, while fine adjustments were performed using the manipulator (\fig{well-positioning}(a)). Subsequently, longitudinal docking of the inflatable modules was demonstrated. For this test, HIDAS was supported by caster wheels to reduce friction under Earth gravity. Two robots were positioned on both sides of HIDAS and operated synchronously to apply pushing forces for alignment (\fig{well-positioning}(b)).

These demonstrations successfully showcased the advantage of the modular and reconfigurable architecture, in which a MoonBot Dragon was separated into two independent Minimal robots that collaboratively achieved precise positioning of the inflatable module. Using a force gauge, the force required to move the half-scale HIDAS was measured as 200 $\pm$ 40\;N for rotation with the ring attachment and 120 $\pm$ 20\;N for dragging with caster wheels, under Earth gravity. In particular, we confirmed the effectiveness of the ring attachment in reducing rolling friction, such that a single Minimal robot was able to rotate the module even under Earth gravity.

\subsection{Safety locking}
\begin{figure}[t]
    \centerline{\includegraphics[width=\linewidth]{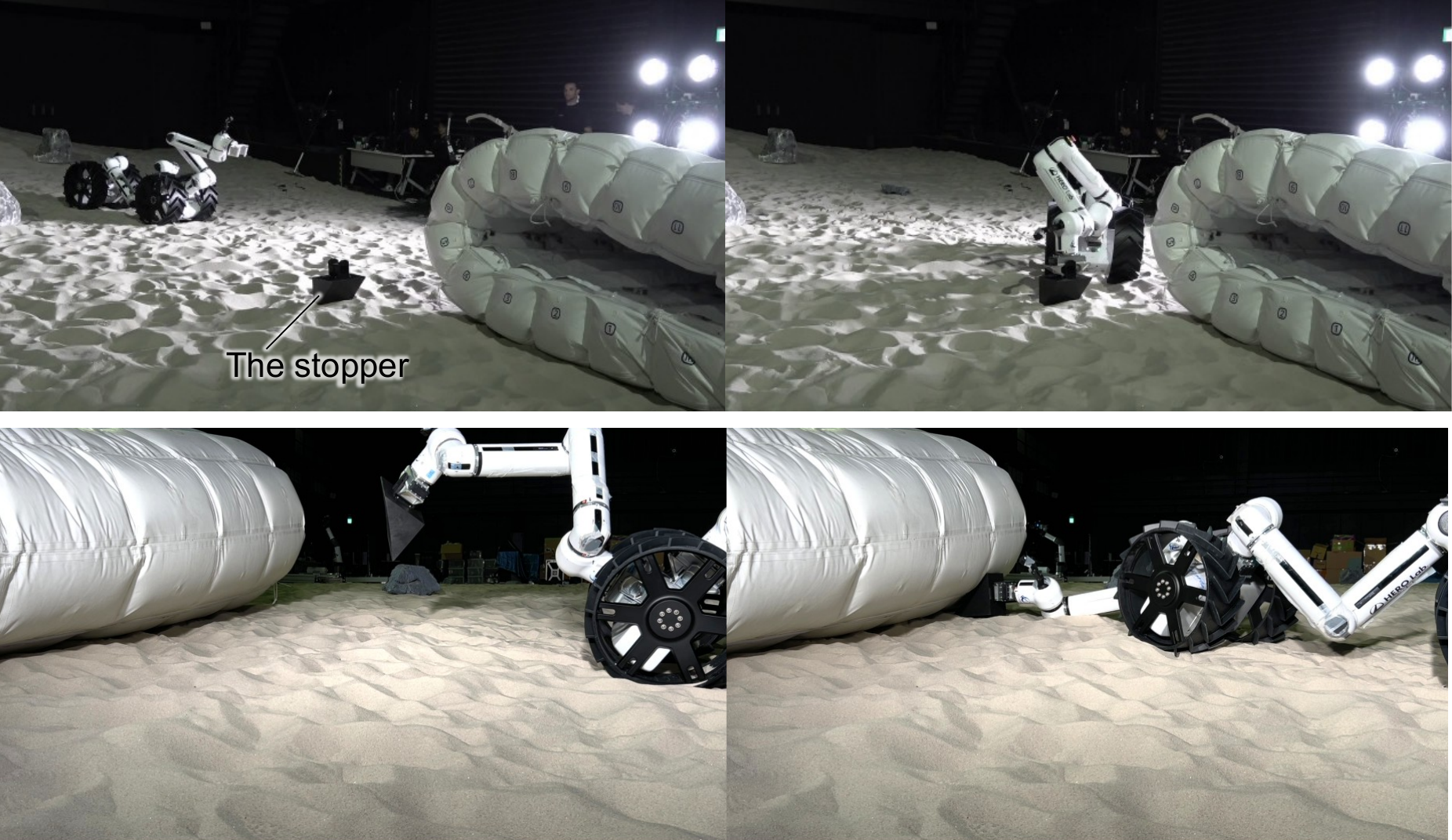}}
    \caption{Demonstration of inflatable module deployment assistance by MoonBot. Once the inflatable module had been fully deployed and properly positioned, the robot placed stopper objects to further enhance stability (top: pick-up, bottom: insertion).
\label{stopper_insertion}}
\end{figure}

Once both inflation and positioning were complete, MoonBot performed a final stabilization task by inserting a stopper. The stopper was designed with a handle compatible with MoonBot's gripper fixture, enabling teleoperated pick-and-place operations. 
In this demonstration, the first version of MoonBot {\it Dragon}, which was also used in the inflation inspection demonstration, was manually teleoperated by the operator. A morphology-specific controller, which incorporates one limb bridging two wheels for steering and the other limb serving as a front manipulator, was used for this configuration.
As a result, the robot successfully positioned the stopper beneath the side of HIDAS to prevent unintended movement (\fig{stopper_insertion}).

This pick-and-place demonstration represents similar tasks, such as attaching additional shielding covers or inserting a docking hub into the inflatable module.

\subsection{Post-deployment 3D reconstruction}
\begin{figure}[t]
    \centerline{\includegraphics[width=\linewidth]{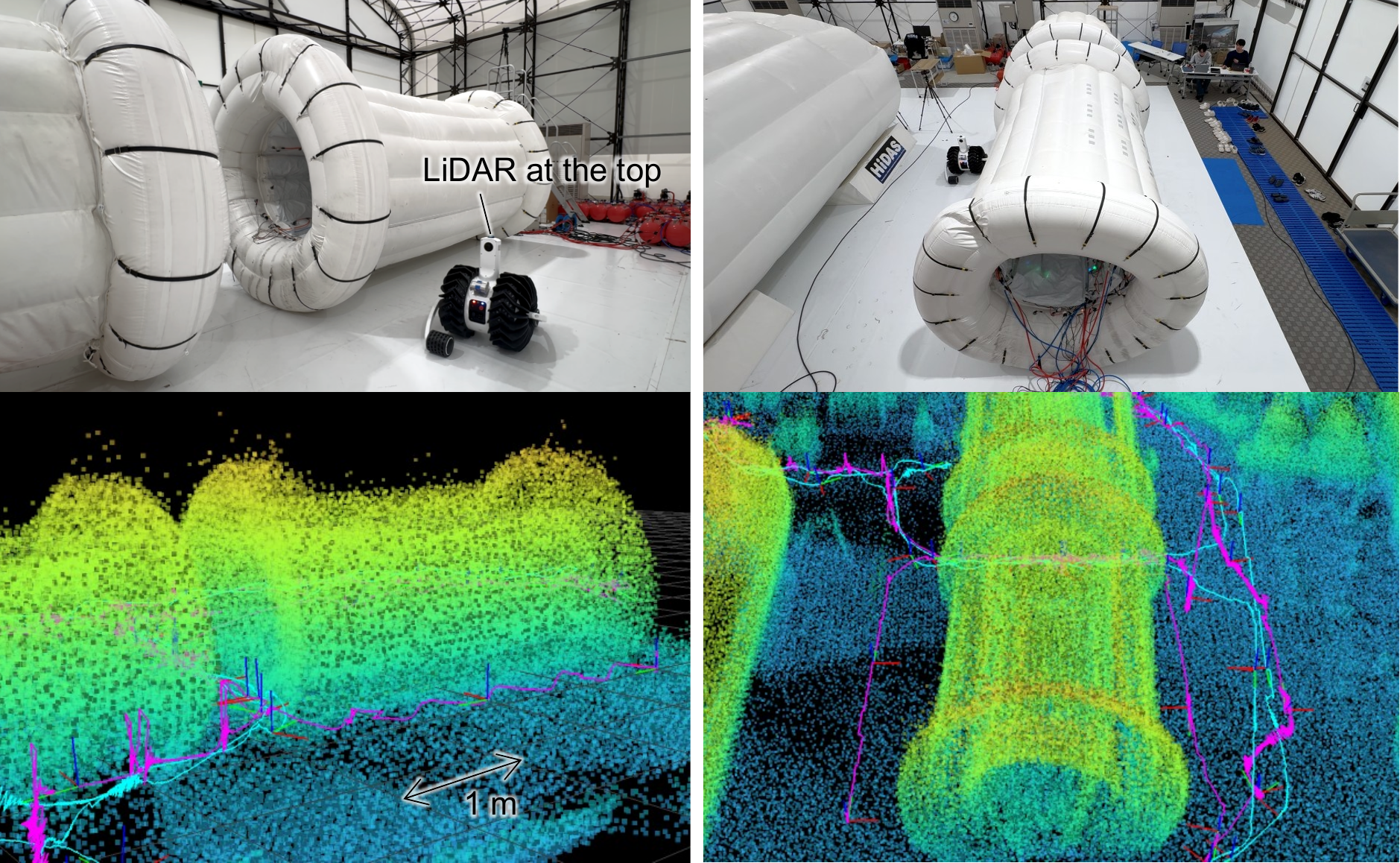}}
    \caption{Three-dimensional reconstruction of the inflatable modules after deployment. Multiple MoonBots scanned the HIDAS modules using LiDAR sensors mounted on top of the wheel bases (top images) to generate point cloud data (bottom graphics). The complete geometry of HIDAS was reconstructed by fusing multiple scans; the magenta and cyan lines indicate the robot traveling trajectories estimated during two different scanning runs.
\label{3Dmapping}}
\end{figure}

As a post-inflation procedure, the robots reconstruct the three-dimensional geometry of the inflated modules. This process is essential for verifying the completion of deployment, while the resulting 3D mapping data are also useful for future maintenance tasks. To demonstrate this, a MoonBot Wheel module equipped with a LiDAR sensor (Livox Mid-360) mounted on top of the base was driven by operators from a hidden location to scan HIDAS. For simultaneous localization and mapping (SLAM), GLIM~\cite{koide2024glim}, a state-of-the-art SLAM framework, was integrated into our software system. Point cloud data acquired from multiple scanning passes were fused, resulting in the successful reconstruction of the full geometry of HIDAS (see \fig{3Dmapping}). The reconstructed geometry of the module can be further utilized as a landmark for precise robot localization and for docking tasks involving additional external components, such as airlock doors and outer shielding shells.

To evaluate the accuracy of the reconstruction, the dimensions of HIDAS were estimated from the reconstructed point cloud. The external diameter and longitudinal length of the main cylindrical body were approximately 2.2\;m and 2.9\;m, respectively. Compared with the nominal dimensions of 2.0\;m and 3.0\;m, these values correspond to deviations of approximately 8\% and 3\%, indicating dimensional agreement within approximately 10\%. 
\subsection{Summary}
Table~\ref{tab:summary} summarizes the overall results of the field demonstrations. Using the developed teleoperation framework, human operators were able to supervise and carefully control the robots remotely, allowing most of the demonstrated tasks to be completed without any physical intervention during operation. Although the system was not fully autonomous, retaining human judgment in the control loop provided operators with greater confidence when performing contact-rich and safety-critical tasks, particularly when unexpected conditions arose during operation (e.g., a sudden communication interruption).
\begin{table}[b]
  \centering
  \footnotesize 
  \caption{Summary of the field demonstrations.}
  \vspace{-2mm}
  \label{tab:summary}
  \setlength{\tabcolsep}{3pt}
  \begin{tabular}{l|rrr}
  \toprule
  Demonstration & Trials & Successes & Operation time [min]\\ 
  \midrule
  Inflation inspection    & 4 & 4$^{~\;\;}$ & 20, 10.5, 1.5$^{1)}$, 1.5$^{1)}$  \\
  Lateral well-positioning  & 3 & 2$^{2)}$ & 3, 2, 1.5  \\
  Longitudinal well-positioning  & 3 & 2$^{2)}$ & 4, 3, 2.5 \\
  Safety stopper pick-and-place & 3 & 3$^{~\;\;}$ & 16, 15, 7 \\
  3D reconstruction & 1 & 1$^{~\;\;}$ & 26.5  \\
  \bottomrule
  \end{tabular} \\
  \raggedright \scriptsize
  \vspace{.5mm}
  ~$^{1)}$The contact-based inspection process was omitted in the latter two trials because no anomalous cells were detected during the visual inspection.\\ 
  ~$^{2)}$These two trials were judged unsuccessful because the intended final relative alignment was not reached, as determined by the operator from the live camera views.
\end{table}

The operation time was also found to depend strongly on operator proficiency. As shown in Table~\ref{tab:summary}, the time required to complete several tasks decreased substantially over repeated trials. This reduction was mainly attributed to the operators becoming increasingly familiar with the robot behavior, camera views, and teleoperation interface. 

\section{Lessons Learned}
The primary objective of this study is not only to demonstrate the feasibility of robot-assisted deployment of inflatable habitat modules but also to extract practical lessons that can guide future lunar missions. Based on the field demonstrations, several practical lessons were identified regarding robotic inspection, manipulation, positioning, and system operation.

First, during the inflation inspection, we found that vision-based anomaly detection is highly sensitive to illumination conditions. Although the custom-trained model successfully detected under-inflated cells in the reproduced lunar lighting environment, reliable performance requires sufficiently realistic and more diverse training data. This lesson suggests that photorealistic virtual datasets are highly desirable for practical robotic inspection of inflatable modules on the Moon. Since acquiring a large number of real lunar images is impractical, a practical strategy would be to train the model using large-scale synthetic datasets generated from a digital twin of the lunar environment~\cite{omnilrs} together with realistic hardware models~\cite{hidas_sim}, and subsequently fine-tune the model using a limited number of real images acquired on the lunar surface.

Second, the well-positioning demonstrations revealed a clear difference between rotational repositioning and longitudinal transportation of the inflatable module. While the ring attachment effectively reduced rolling resistance, allowing a single MoonBot Minimal to rotate the half-scale HIDAS even under Earth gravity, longitudinal dragging remained considerably more demanding because of the large contact area between the module and the ground. This observation indicates that robot-assisted rotational positioning is a practical and effective operation, whereas longitudinal transportation requires additional mobility support from the inflatable module itself or increased robotic traction. These findings suggest that the robot and the inflatable module should complement each other's mobility capabilities rather than relying solely on robotic pushing.

Assuming the same rolling friction observed in the demonstrations, the force required to rotate the full-scale HIDAS (diameter: 4\;m), which has 2$^3$ times the mass of the half-scale prototype, is estimated to be approximately 267\;N under lunar gravity (1/6\;G). This value is within the capability of a single MoonBot Minimal configuration; however, it may increase because of higher rolling resistance on lunar regolith. Therefore, combining the self-rotation capability of HIDAS with robotic assistance or reconfiguring MoonBot into a stronger morphology (e.g., \textit{Dragon}) is expected to improve operational robustness. Likewise, longitudinal transportation would benefit from the ``inchworm-like'' self-locomotion capability enabled by HIDAS, which will be investigated experimentally in future work.


Third, the manipulation demonstrations confirmed that robot-assisted pick-and-place operations are a practical approach for handling external components around inflatable habitat modules. Although the present experiments focused on stopper insertion, the same manipulation strategy is directly applicable to many representative assembly tasks required during lunar habitat deployment. These demonstrations indicate that robotic manipulation can effectively reduce direct human intervention during infrastructure construction while maintaining accurate component placement.

The experiments also highlighted the importance of robotic adaptability for handling components with different sizes, masses, and installation locations. Future habitat modules will require additional structures, including pressurized inner shells for human habitation sections, airlock hatches providing access to the inner shell, docking hubs for connecting multiple modules, and external shielding covers. Many of these components must be transported without contaminating them with regolith and installed at elevated positions above the ground. These observations suggest that a modular and reconfigurable robotic system provides significant operational advantages, as it can either reconfigure into high-performance morphologies for heavy manipulation or separate into multiple robotic units that collaboratively perform distributed assembly tasks according to mission requirements, as illustrated in \fig{fig2}.

Lastly, another important lesson learned concerns the appropriate level of autonomy. Throughout the demonstrations, autonomous execution proved sufficiently reliable for the non-contact tasks investigated in this study, including visual recognition of the inflatable module, autonomous approach, and LiDAR-based three-dimensional mapping, where unexpected failures pose relatively little risk to the hardware. In contrast, contact-rich tasks, such as haptic inspection, module positioning by pushing, and component insertion, required continuous human supervision because the contact conditions changed dynamically and even small positioning errors could lead to mission-critical failures. Although these operations could potentially be automated in the future, our field demonstrations showed that human-in-the-loop teleoperation provided substantially greater operational confidence and robustness for these safety-critical tasks. These observations suggest that selecting the level of autonomy according to the operational risk of each task, rather than aiming for full autonomy, is a practical operational principle for future robotic lunar infrastructure construction.

\section{Conclusion}\label{conclusion}
In this paper, as one of the most practical approaches toward the reliable establishment of foundational human habitat modules, robot-assisted deployment and maintenance are demonstrated using actual hardware, MoonBot and HIDAS, in lunar analogue environments. The proposed combined system, consisting of a modular and reconfigurable robot and inflatable modules, functions as a unified heterogeneous modular system. The operation software developed for the demonstration facilitates human-in-the-loop teleoperation for robotic loco-manipulation tasks.

Throughout the field demonstrations, the robot successfully performed representative deployment tasks, including haptic inspection during inflation, module positioning and alignment, and pick-and-place operations of external components. We also demonstrated vision-enhanced maintenance functions, including image-based anomaly detection and post-deployment three-dimensional reconstruction. The results confirmed the sufficient loco-manipulation capability of MoonBot (applicable force exceeding 100\;N per Minimal robot) and the effectiveness of coordinated operation by multiple robots through wireless teleoperation, and provided practical insights into robot-assisted deployment strategies for inflatable habitat modules.

This work concludes that robotic assistance through human-supervised teleoperation is an effective approach for the secure and reliable deployment and maintenance of inflatable modules toward lunar outpost construction. Furthermore, the lessons learned from these demonstrations establish practical guidelines for future robot-assisted deployment systems, thereby contributing to the advancement of lunar habitat technologies.

\section*{Acknowledgment}
The authors would like to thank the Hyper-Environmental Robots Laboratory (HERO Lab.) at Hakusan Corporation, Tokyo, Japan, for their invaluable support in the development of MoonBot, and Asteria ART, ISP, and JAOPS for their effective support in software development.
The authors also acknowledge the members of Prof. Shinichi Kimura's laboratory at Tokyo University of Science for their operation of HIDAS during the collaborative experiments with MoonBot.
The authors are also grateful to the staff of the Advanced Facility for Space Exploration at JAXA Sagamihara campus for their kind support in the field testing.
Finally, the authors would like to thank the SRL members for their continuous assistance through the field test campaigns.

\bibliographystyle{IEEEtran}
\bibliography{reference.bib}

\end{document}